\documentclass[fleqn,10pt]{wlscirep}
\usepackage[utf8]{inputenc}
\usepackage[T1]{fontenc}
\usepackage{cite}
\usepackage{amsmath,amssymb,amsfonts}
\usepackage{algorithm}
\usepackage{algorithmic}
\usepackage{graphicx}
\usepackage{textcomp}
\usepackage{xcolor}
\title{Machine Learning-powered Compact Modeling of Stochastic Electronic Devices using Mixture Density Networks}

\author[1]{Jack Hutchins}
\author[1]{Shamiul Alam}
\author[2]{Dana S. Rampini}
\author[2]{Bakhrom G. Oripov}
\author[2]{Adam N. McCaughan}
\author[1,*]{Ahmedullah Aziz}
\affil[1]{Department of Electrical Eng. \& Computer Sci., University of Tennessee, Knoxville, Tennessee 37996, USA}
\affil[2]{National Institute of Standards and Technology, Boulder, Colorado 80305, USA}

\affil[*]{Corresponding Author. Email: aziz@utk.edu }

\begin{abstract}
The relentless pursuit of miniaturization and performance enhancement in electronic devices has led to a fundamental challenge in the field of circuit design and simulation—how to accurately account for the inherent stochastic nature of certain devices. While conventional deterministic models have served as indispensable tools for circuit designers, they fall short when it comes to capturing the subtle yet critical variability exhibited by many electronic components. In this paper, we present an innovative approach that transcends the limitations of traditional modeling techniques by harnessing the power of machine learning, specifically Mixture Density Networks (MDNs), to faithfully represent and simulate the stochastic behavior of electronic devices. We demonstrate our approach to model heater cryotrons, where the model is able to capture the stochastic switching dynamics observed in the experiment. Our model shows 0.82\% mean absolute error for switching probability. This paper marks a significant step forward in the quest for accurate and versatile compact models, poised to drive innovation in the realm of electronic circuits.
\end{abstract}
\begin{document}

\flushbottom
\maketitle
%
%
\thispagestyle{empty}

\section*{Introduction}

In the ever-evolving landscape of electronic devices, the pursuit of smaller, faster, and more efficient technology has led to groundbreaking innovations, unlocking new possibilities for various applications across industries \cite{salahuddin2018era}. However, as devices continue to shrink and push the boundaries of Moore's law \cite{lundstrom2003moore}, the inherent stochastic nature of certain electronic components has become increasingly significant \cite{alawad2016survey}. These stochastic electronic devices exhibit inherent variability in their behavior, posing a formidable challenge for conventional deterministic modeling approaches \cite{hamilton2014stochastic}.

In the pursuit of greater accuracy and ease of implementation in electronic circuit simulations, a significant leap has been taken by incorporating machine learning methodologies into compact modeling \cite{Li2016, Zhang2017, Wang2021, Abouelyazid2022, Lin2022, Hutchins2022}. While various machine learning-based compact models have emerged, these models, thus far, have failed to address the implications of stochastic behavior in electronic devices. And while traditional physics-based compact models have been created with stochastic properties\cite{Wang2014, Becle2021}, these models lack the benefits of machine learning-based methods such as decreased turnaround time and little required device knowledge. The dynamics of electronic devices, traditionally considered deterministic, are increasingly proving to be intrinsically stochastic at the nanoscale. Consequently, conventional compact models that ignore these inherent stochastic processes fail to capture the true behavior of the devices, leading to inaccuracies in circuit simulation.

This paper presents a new approach that addresses this gap in the realm of compact modeling. We introduce the use of Mixture Density Networks (MDNs)\cite{Bishop1994} to capture the stochastic nature of electronic devices accurately. In addition to this neural network approach, we introduce a novel sampling methodology that uses inverse transform sampling \cite{Devroye1986} to make our model more accurate to the stochastic nature of devices. By doing these, we create compact models that account for the inherent stochasticity of certain devices, allowing the development of new circuits that are both reliable and innovative. Our approach represents a paradigm shift in electronic device modeling, as it for the first time encompasses the full spectrum of electronic device behavior, from deterministic to stochastic.

To demonstrate the effectiveness and versatility of our approach, we focus on the modeling of heater cryotron \cite{mccaughan2014superconducting}, a three-terminal device that shows gate-current controlled switching between superconducting and resistive states. Heater cryotrons, with their stochastic switching behavior, are a particularly compelling case study. Our methodology accurately captures the nuances of this stochasticity, providing a foundation for the development of novel circuits and the optimization of existing designs.

The contributions of this paper are threefold:

\begin{itemize}
    \item \textbf{Mixture Density Networks}: By employing MDNs, we equip our modeling framework with the capacity to capture complex probability distributions, thereby enabling a more faithful representation of device variability. This empowers designers to explore device behavior beyond traditional deterministic bounds.
    \item \textbf{Inverse Transform Sampling}: By leveraging inverse transform sampling, we can use our MDN to its full potential. This approach allows the model to output smooth outputs in transient simulations while maintaining stochasticity. This in turn makes our model more realistic to device-to-device or cycle-to-cycle variation.
    \item \textbf{Demonstration with Heater Cryotrons}: To showcase the effectiveness of our approach, we focus on modeling heater cryotrons—a class of devices known for their stochastic switching behavior. Through this demonstration, we illustrate the practical applicability of our methodology in real-world electronic systems.
\end{itemize}

As we delve into the details of our approach, we will elucidate the intricacies of MDN-based modeling for stochastic electronic devices, providing a comprehensive understanding of the benefits it offers to circuit simulation and electronic design. 

\section*{Results}
The following section will describe our neural network architecture, sampling methodology, and the results of our approach using experimentally derived heater cryotron data.
\subsection*{Mixture Density Network}

Mixture density networks provide the perfect approach for modeling stochastic devices. Mixture density networks work similarly to standard multilayer perceptions (MLP)\cite{Hornik1989}, but with three output layers connected to the last hidden layer instead of just one. Each output layer has the same number of neurons $N$, and we label the output layers as $\mu$, $\sigma$, and $\alpha$. The output of these layers is used as the parameters of a Gaussian mixture distribution probability density function (PDF) as defined in equation (1).

\begin{equation}
p(x) = \sum_{k=1}^{N} \alpha_k \cdot \frac{1}{\sqrt{2\pi\sigma_k^2}} \cdot \exp\left(-\frac{(x - \mu_k)^2}{2\sigma_k^2}\right)
\end{equation}

Using this approach allows the model to learn the unique output distribution for any given input. The PDF is a combination of $N$ different Gaussian distributions, which are multiplied by a scaling factor $\alpha$. In order to maintain a valid probability distribution, we need to ensure that the sum of $\alpha$ is always equal to 1. To do that, we apply a softmax function to the output of the $\alpha$ layer, as defined in equation (2).

\begin{equation}
    s(\alpha_i) = \frac{e^{\alpha_i}}{\sum_{j=1}^N e^{\alpha_j}}
\end{equation}

Additionally, we need to ensure that the output of the $\sigma$ layer needs to be positive and non-zero since they represent the standard deviation of the Gaussian distributions in the mixture density PDF. In order to do this, we use the exponential linear unit (ELU) activation function plus 1 plus $\epsilon$, where $\epsilon$ is the smallest value before loss of precision in floating point calculations. The activation function can be seen in equation (3).

\begin{equation}
ELU(\sigma_i) = \left\{
    \begin{array}{lr}
        \sigma_i + 1 + \epsilon, & \text{if } \sigma \geq 0\\
        0.5(e^{\sigma_i} - 1) + 1 + \epsilon, & \text{if } \sigma < 0
    \end{array}
\right\}
\end{equation}

Finally, we allow $\mu$ to be any value, so there is no need to use an activation function to restrict its output. For the hidden layers, we chose to use the rectified linear unit activation function, which is defined in equation (4).

\begin{equation}
ReLU(x) = \left\{
    \begin{array}{lr}
        x, & \text{if } x \geq 0\\
        0, & \text{if } x < 0
    \end{array}
\right\}
\end{equation}

In order to train the model, we need an appropriate loss function to measure the performance of the model. We choose to use Gaussian Negative Log-Likelihood (GNLL), since it effectively captures the performance with respect to probability. The GNLL for a single data point $x$ is the negative logarithm of this PDF:

\begin{equation}
\text{NLL}(x) = -\log\left(\sum_{k=1}^{N} \alpha_k \cdot \frac{1}{\sqrt{2\pi\sigma_k^2}} \cdot \exp\left(-\frac{(x - \mu_k)^2}{2\sigma_k^2}\right)\right)
\end{equation}

The goal during training is to minimize the GNLL across all training data points. Therefore, the overall GNLL loss is the average GNLL over all data points:

\begin{equation}
L = -\frac{1}{N} \sum_{i=1}^{N} \log\left(\sum_{k=1}^{K} \alpha_k \cdot \frac{1}{\sqrt{2\pi\sigma_k^2}} \cdot \exp\left(-\frac{(x_i - \mu_k)^2}{2\sigma_k^2}\right)\right)
\end{equation}

\begin{figure*}[tb!]
    \centering
    \includegraphics[width=17cm]{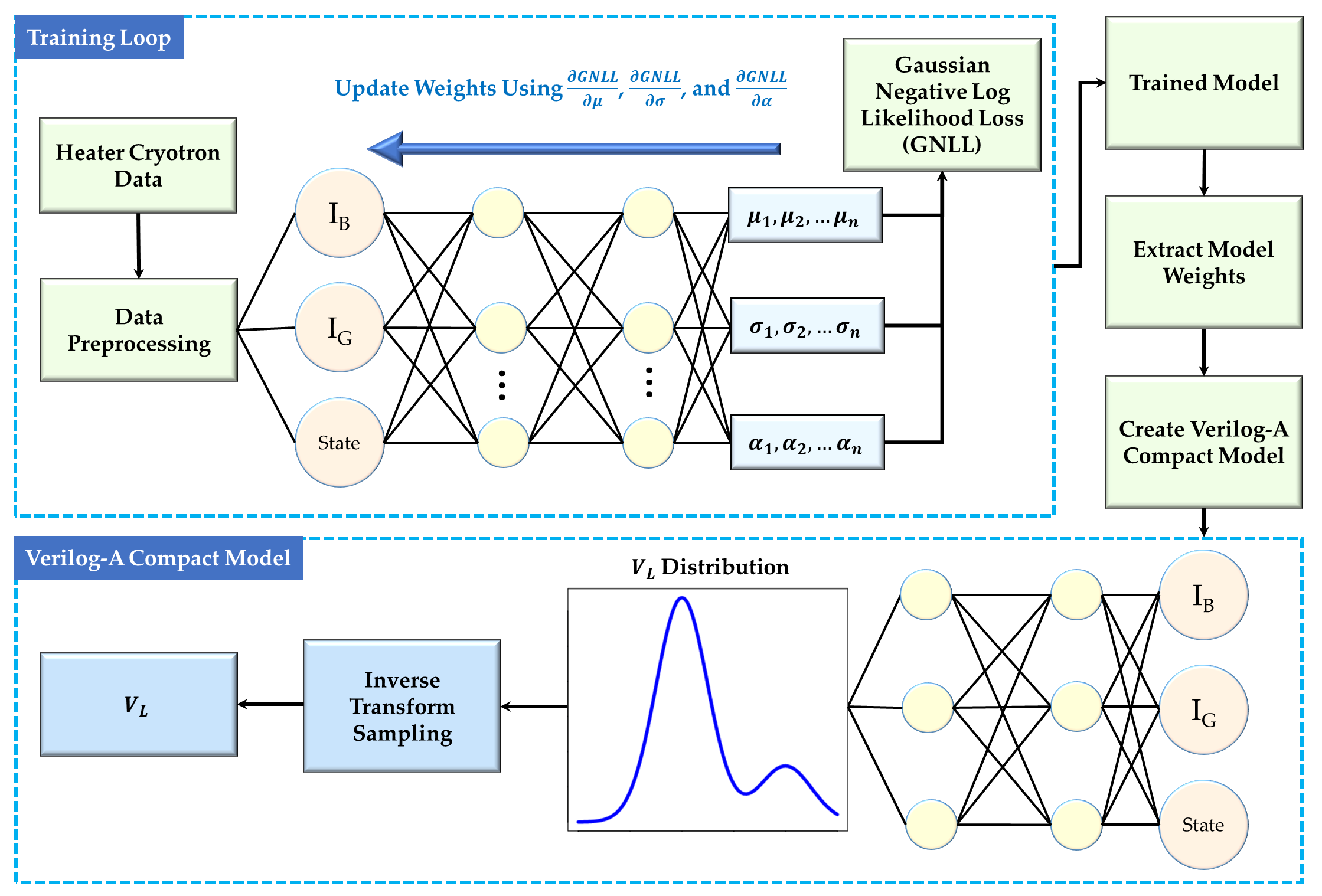}
    \caption{\label{fig:1} \textbf {Overview of the machine learning-powered modeling approach for the stochastic behavior of heater cryotron.} After training the mixture density network with the device characteristics obtained from experiments, the weights are extracted which are eventually used to create the Verilog-A-based compact model for circuit/system-level simulations. }
\end{figure*}

During training, we adjust the network's parameters to minimize this GNLL loss. We can use backpropagation and the AdamW\cite{loshchilov2019decoupled} optimizer to update the network's weights and biases by finding the gradients of GNLL with respect to $\mu$, $\sigma$, and $\alpha$ as shown in figure \ref{fig:1}.

\subsection*{Model Sampling Methodology}
\begin{algorithm}[tbp]
\caption{Standard Sampling from a Gaussian Mixture Density Distribution}
\label{alg:gaussian_mixture_sampling}
\begin{algorithmic}
\STATE \textbf{Input}:
\begin{ALC@g}
\STATE $K$ - Number of Gaussian components
\STATE $\{\mu_1, \mu_2, \ldots, \mu_K\}$ - Means of the Gaussians
\STATE $\{\sigma_1, \sigma_2, \ldots, \sigma_K\}$ - Stand deviation of the Gaussians
\STATE $\{\alpha_1, \alpha_1, \ldots, \alpha_1\}$ - Mixing coefficients, $\sum_{i=1}^{K} \pi_i = 1$
\end{ALC@g}
\STATE \textbf{Output}:
\begin{ALC@g}
\STATE Sample $x$ from the mixture distribution
\end{ALC@g}
\STATE \textbf{Sample Function}:
\begin{ALC@g}
\STATE sum = 0
\STATE Sample $q$ from a uniform distribution [0, 1]
\STATE \textbf{for} $k \in \{1,\ K\}$ \textbf{do}
\begin{ALC@g}
\STATE sum = sum + $\alpha_k$
\STATE \textbf{if} sum $\geq\ q$ \textbf{then}
\begin{ALC@g}
\STATE\textbf{break}
\end{ALC@g}
\end{ALC@g}
\STATE \textbf{end}
\STATE Sample $x$ from the Gaussian distribution $\mathcal{N}(\mu_k, \sigma_k)$
\end{ALC@g}
\STATE \textbf{Return}:
\begin{ALC@g}
\STATE $x$ - Sample from the Gaussian mixture density
\end{ALC@g}
\end{algorithmic}
\end{algorithm}

In order to properly utilize our model, it is critical that we use an appropriate sampling methodology for the output distributions. The obvious approach for this would be to randomly sample from the distribution, however, this approach presents several issues. In order to sample in this way, we can use the approach outlined in Algorithm \ref{alg:gaussian_mixture_sampling}, where we randomly select a distribution $k$ for the mixture density distribution and sample from a standard normal distribution $\mathcal{N}(\mu_k, \sigma_k)$. This approach allows us to utilize uniform and standard Gaussian sampling, which is built into most modeling languages including Verilog-A. The issue with this approach is that it results in sporadic currents in transient simulations since the current will jump around the distribution every time step. In addition, this approach will cause multi-state devices to switch sporadically between states near the critical value if the switching of the device is stochastic. While this approach may be valuable for some devices, for most devices we will need another sampling methodology that more accurately reflects the variation we see in devices. We propose using inverse transform sampling\cite{Devroye1986} to accomplish this. 

Inverse transform sampling works by first obtaining the cumulative distribution function (CDF) of the desired distribution and then finding its inverse. A cumulative distribution function (CDF) is a function that gives the probability that a random variable is less than or equal to a specific value. In the context of inverse transform sampling, the input (a uniform random number between 0 and 1) represents a quantile of the distribution because it corresponds to a specific probability level, and the inverse CDF maps it to the corresponding value in the distribution. To find the CDF, we can start with the probability density function in equation (7).

\begin{equation}
p(x) = \sum_{k=1}^{K} \alpha_k \cdot \frac{1}{\sqrt{2\pi\sigma_k^2}} \cdot \exp\left(-\frac{(x - \mu_k)^2}{2\sigma_k^2}\right)
\end{equation}

Our CDF will be equal to the integral from $-\infty$ to $x$ as shown in equation (8).

\begin{equation}
F(x) = \int_{- \infty}^{x} \sum_{k=1}^{K} \alpha_k \cdot \frac{1}{\sqrt{2\pi\sigma_k^2}} \cdot \exp\left(-\frac{(t - \mu_k)^2}{2\sigma_k^2}\right) dt
\end{equation}

While this is the CDF, it would be beneficial for us to manipulate this into a more calculable form. By bringing the integral inside the summation and leaving the scaling factor $\alpha$ outside the integral, we are left with equation 9.

\begin{equation}
F(x) = \sum_{k=1}^{K} \alpha_k \int_{- \infty}^{x} \frac{1}{\sqrt{2\pi\sigma_k^2}} \cdot \exp\left(-\frac{(t - \mu_k)^2}{2\sigma_k^2}\right) dt
\end{equation}

This leaves us with the weighted sum of the CDF of a standard normal distribution, which can be defined in terms of the error function as shown in equation (10).

\begin{equation}
F(x) = \sum_{k=1}^{K} \frac{\alpha_k}{2}\left[1+erf\left(\frac{x-\mu_k}{\sigma_k \sqrt{2}}\right)\right]
\end{equation}

Defining this in terms of the error function makes calculation much easier since the error function is well-defined and easy to calculate numerically. The error function is defined in equation (11). 

\begin{equation}
erf(x)=\frac{2}{\sqrt{\pi}}\int_{0}^{x}e^{-t^{2}}\, dt
\end{equation}

The issue is that there is no closed-form solution to the inverse of this CDF. As such, we use a numerical root finder to find what value of x leads to $F(x)- q=0$. We choose to use Brent's method\cite{Brent1973}, however, most numerical root finders should work here. For derivative-based root finders, the derivative of the CDF is equal to the PDF.

Now that we have established a way to perform inverse transform sampling, we can take full advantage of its properties to improve our model. To solve our issue with transient simulations, we can keep the same value of $q$ for the entirety of a sweep. This will result in a continuous output that is more in line with the device's properties of cycle-to-cycle variation than traditional sampling. In this case, we can generate a new $q$ every time the derivative of an input with respect to time crosses 0, which can easily be done in Verilog-A. Another benefit of this sampling approach is the ability to clip the probability distribution. For example, we could generate $q$ only between 0.05 and 0.95 so that the model doesn't predict values more than 2 standard deviations away (or in any range). Another use could be modeling device-to-device variation by weighting the distribution when sampling for $q$, though this work doesn't explore this option.

\subsection*{Device Characteristics of Heater Cryotron}
\floatstyle{plain}
\restylefloat{figure}
\begin{figure*}[tb!]
    \centering
    \includegraphics[width=17cm,trim=4 40 4 4,clip]{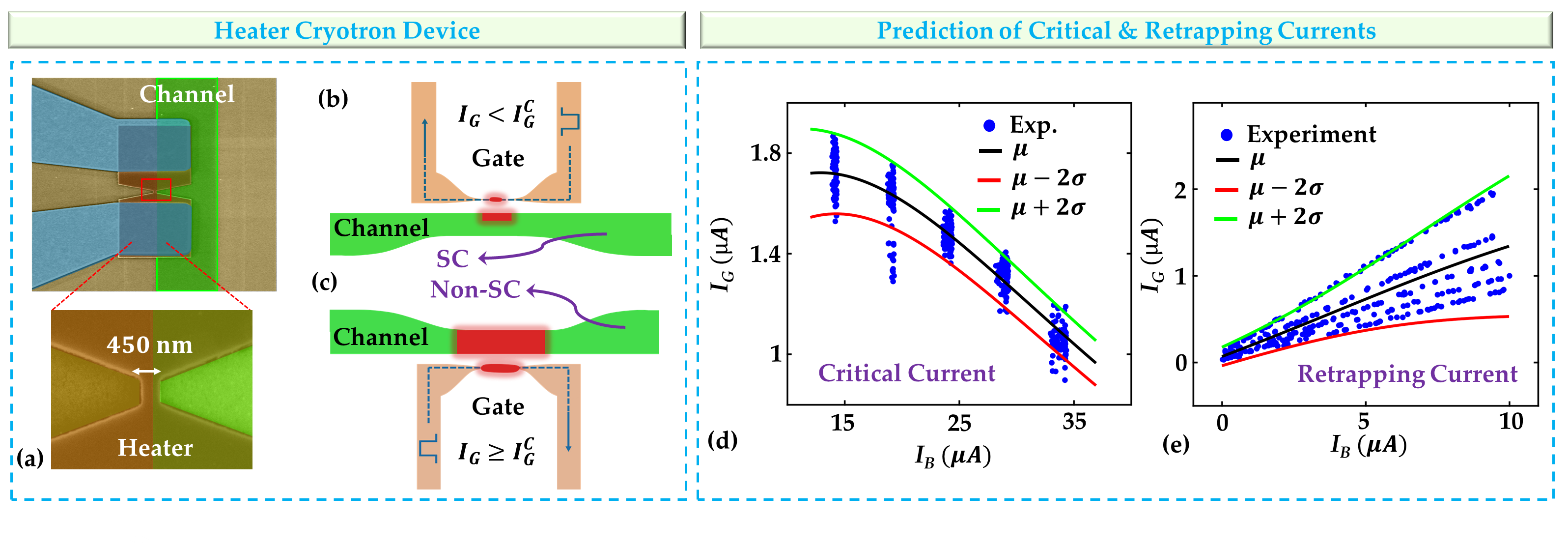}
    \caption{\label{fig:2} \textbf{Prediction of Switching characteristics of heater cryotron.} (a) A false-colored scanning electron microscope (SEM) image of heater cryotron device consisting two superconducting nanowires of tungsten silicide ($WSi$) and a dielectric spacer of $SiO_2$. (b)-(c) Illustration of the gate-current-controlled switching mechanism. An enough gate current ($I_G$) switches the gate nanowire to its resistive state and generates thermal phonos which eventually causes the switching of the channel from its superconducting to resistive state. (d) Switching model for critical current evaluated at the mean ($\mu$) and plus or minus 2 standard deviations ($\sigma$) of the resultant distribution and (e) retrapping current of heater cryotron switching model evaluated at the same points.}
\end{figure*}

The heater cryotron is a superconducting device designed to exploit the unique properties of superconducting nanowires \cite{mccaughan2014superconducting}. This device consists of two superconducting nanowires separated by a dielectric, forming the gate and the channel of the device. Figure \ref{fig:2} (a) shows a false-colored scanning electron microscope (SEM) image of the fabricated heater cryotron device, with a channel width of 1$\mu m$ and a gate width 450$nm$. The gate and channel nanowires are separated by a SiO$_2$ dielectric spacer of 25$nm$ thickness. Initially, for a given channel current ($I_B$), both nanowires (gate and channel), when maintained at cryogenic temperatures, remain superconducting exhibiting zero resistance, and remain so until the gate nanowire becomes resistive (Figure \ref{fig:2} (b)). However, when a sufficient amount of current ($I_G$) is passed through the gate nanowire to make it resistive, it starts generating thermal phonons which transport to the channel nanowire with the help of the spacer. These thermal phonons suppress the superconductivity and reduce the critical current of the channel nanowire ($I_{Ch}^C$). Eventually, when the increase of gate current, reduces the channel critical current below the applied channel current, the channel transitions from its superconducting state to a resistive state as shown in Figure \ref{fig:2} (c). Conversely, removing the current from the heater enables the channel nanowire to revert to its superconducting state. The specific current levels at which superconducting to resistive and resistive to superconducting transitions occur are referred to as the critical current and retrapping current, respectively. The gate current-controlled switching of heater cryotrons has been used in a multitude of applications, including designing logic circuits \cite{mccaughan2014superconducting, alam2022voltage, alam2023cryogenicLogic, alam2023reconfigurable}, memory systems \cite{alam2023cryogenic_review,alam2021cryogenic, alam2022cryogenic, alam2022superconducting, alam2023cryogenic}, and neuromorphic systems \cite{islam2023review,islam2022dynamically, islam2023deep, islam2022design, islam2023cryogenic} for cryogenic environment, and in interfacing superconducting circuits with semiconducting technology \cite{mccaughan2019superconducting}. 

Notably, the point at which the device switches between its superconducting and resistive states is characterized by stochastic behavior, which introduces an inherent element of randomness into its operation. This stochasticity makes this device ideal for testing our MDN-based compact modeling approach. The critical current value is not a fixed constant but instead depends on the applied channel current. For example, a higher bias current leads to a lower critical current for the gate, thus influencing the switching behavior of the device. Conversely, a higher bias current leads to a lower retrapping current. Additionally, a much lower bias current is required to switch from a resistive to a superconducting state since the nanowire produces its own heat when resistive among other factors. Understanding this dependence on the channel current is pivotal in characterizing and modeling the behavior of the heater cryotron.

To sample the characteristics of the device, we sweep the gate at different bias currents. By performing each of the sweeps multiple times and recording the resulting current-voltage characteristics of the device, we can have enough data for the MDN to learn the stochastic switching behavior of the device. This experimental approach allows us to explore the device's response under varying conditions and analyze the interplay between the gate and channel currents.

\subsection*{Heater Cryotron Model}
For our first iteration of the model, we chose to use our MDN to model the critical gate current for a given bias current. This approach allows the use of a simpler neural network structure for our MDN when compared to learning the I-V characteristics directly. To accomplish this, we use bias current and current state of the device as input with the MDN trained to predict gate critical current (if the heater cryotron is in its superconducting state) and retrapping current if the heater cryotron is in its resistive state. The results of this model can be seen in Figure \ref{fig:2} (d) and (e), where we show the experimental data as well as our model's output distribution at the mean ($\mu$), and with a variation of twice the standard deviation ($\sigma$) in either side ($\mu-2\sigma$ and $\mu+2\sigma$) of the mean. We can see that the model is able to capture the critical and recapture current at different bias currents as well as the variance in these switching points. Another benefit to this approach is that in addition to reduced model complexity, we don't have to run the model every time step as long as the bias current is constant, making the model even more efficient in certain applications.

\floatstyle{plain}
\restylefloat{figure}
\begin{figure*}[b!]
    \centering
    \includegraphics[width=17cm,trim=3 10 3 15,clip]{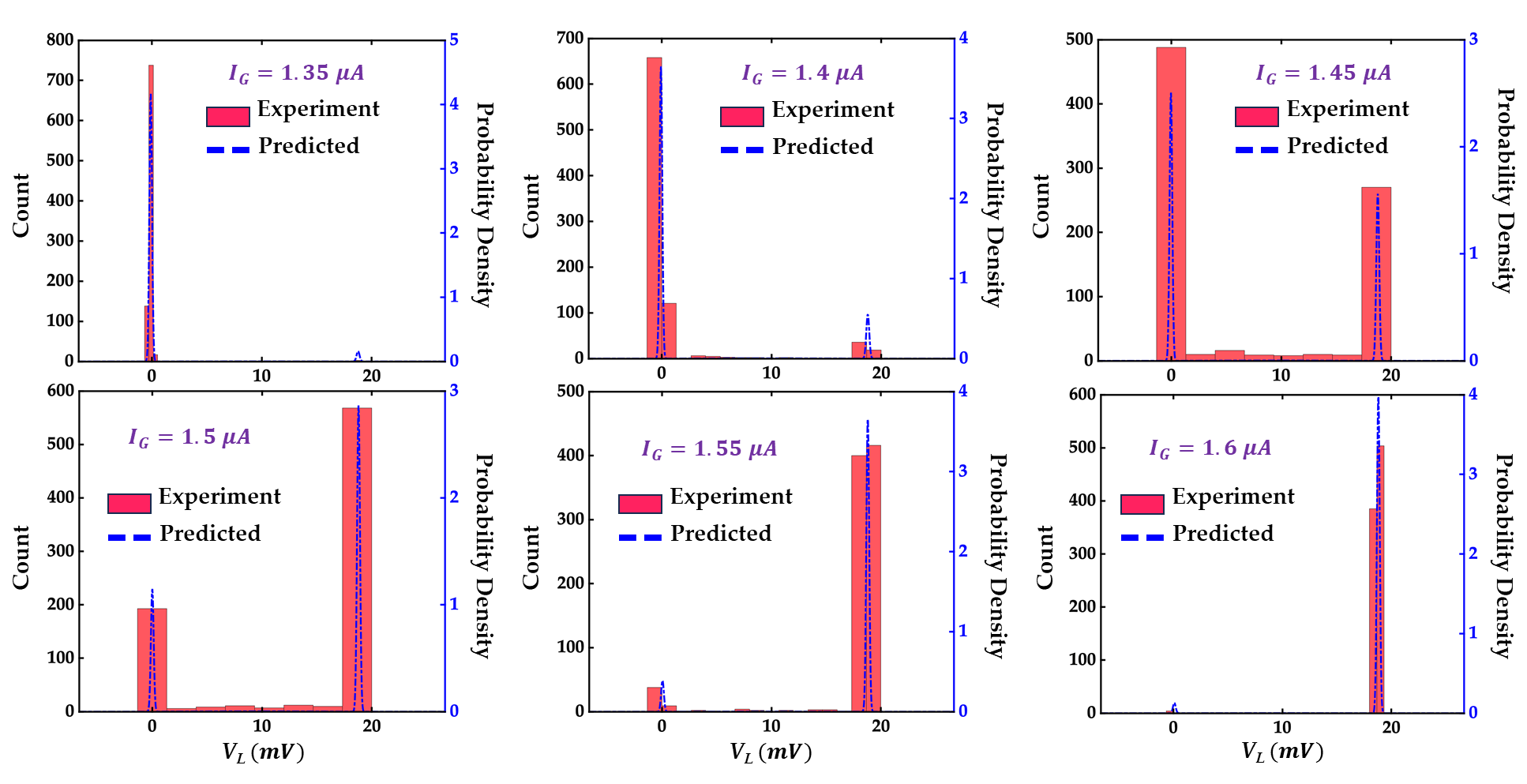}
    \caption{\label{fig:3} \textbf{Model validation for load voltage ($V_L$) against the experiment.} The predicted vs experimental distributions of $V_L$ (dropped across a 1 $k\Omega$ load resistor) at a bias current of 23.5 $\mu A$ and gate current of (a) 1.35 $\mu A$, (b) 1.4 $\mu A$, (c) 1.45 $\mu A$, (d) 1.5 $\mu A$, (e) 1.55 $\mu A$, and (f) 1.6 $\mu A$.}
\end{figure*}

While the switching model could work well for some devices, it comes with some major drawbacks. Most notably, it requires separate models for the I-V characteristics of the device in its different states in addition to the switching model. This means that this approach is unable to capture the variance in the device's I-V characteristics outside of switching. Going forward, we are going to use the MDN to directly learn the I-V characteristics of the heater cryotron. This means that we will be using $I_G$, $I_B$, and state as input to the model, and directly predicting $V_L$ as the output. 

One challenge with MDNs is that there is no good way to provide easily interpretable numerical metrics for performance. Log-likelihood can be used to compare different models, but it is not a useful metric on its own. Since there are not any existing approaches to compare against, this metric will not be useful for us. Because of this, we will need to rely on graphical comparisons between the model and the experimental data. First, we can look at a comparison between the distribution from the model vs a histogram of the experimental data for different values of $I_G$ and $I_B$. This is shown in Figure \ref{fig:3}, where we are using $I_B$ of 23.5$\mu A$ and values of $I_G$ from 1.35$\mu A$ to 1.6$\mu A$. Using this range of $I_G$ allows us to see how the switching probability distributions change around the critical current. Figure \ref{fig:3} shows that our probability distributions closely match the distributions obtained from the experiment.

\begin{figure*}[tb!]
    \centering
    \includegraphics[width=16cm]{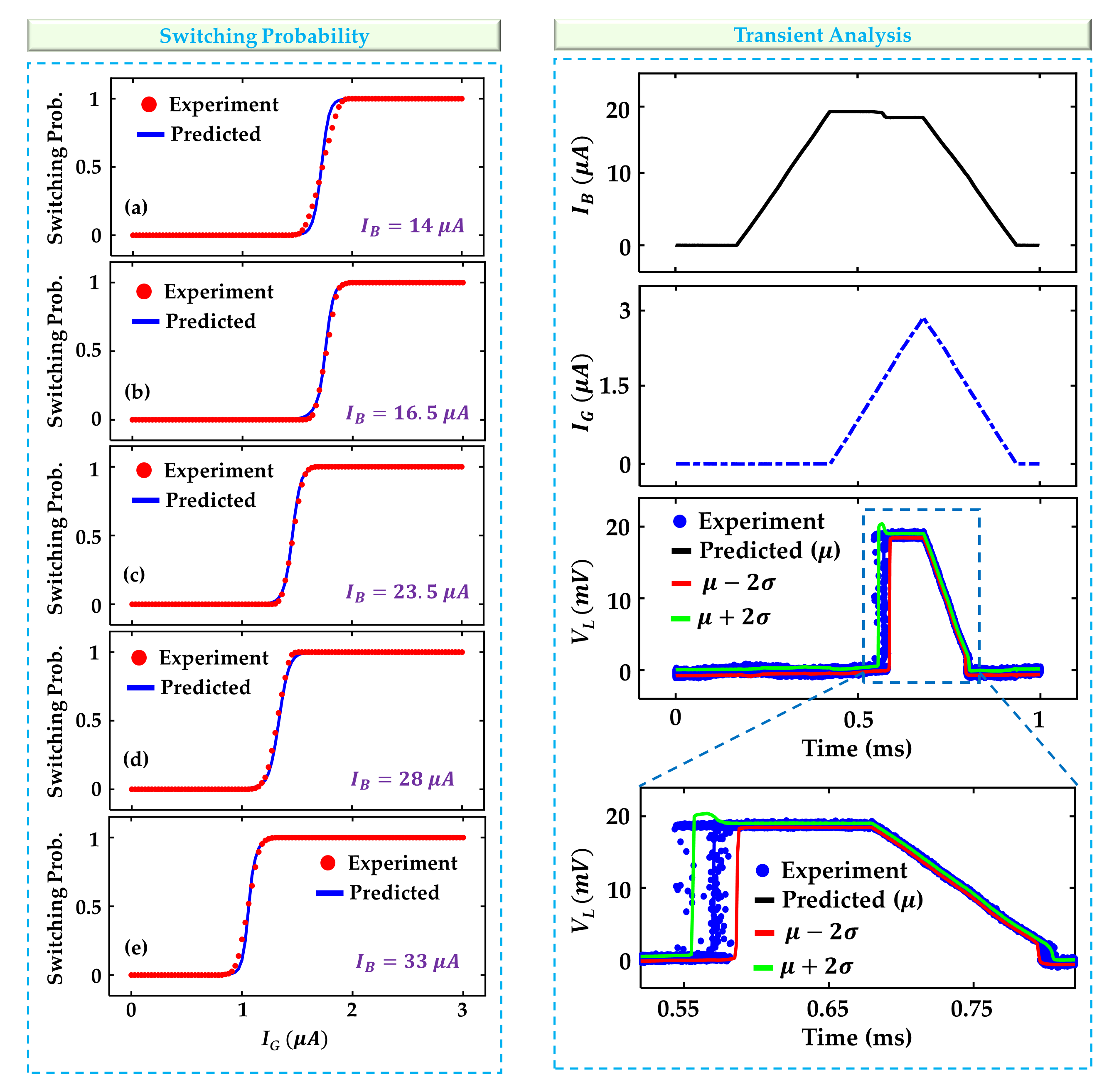}
    \caption{\label{fig:4} \textbf{Switching probabilities at varying bias currents and transient simulation results.} The predicted vs experimental switching probability for gate current ($I_G$) between 0 $\mu A$ and 3 $\mu A$ at a bias current ($I_B$) of (a) 14 $\mu A$, (b) 16.5 $\mu A$, (c) 23.5 $\mu A$, (d) 28 $\mu A$, and (e) 33 $\mu A$. The increase in $I_B$ leads to a reduction in the gate switching current. Transient dynamics of (f) bias current and (g) gate current which cause switching of the channel after exceeding the corresponding thresholds. (h) Switching is shown and validated with the experiment by the time dynamics of the load voltage ($V_L$) at mean ($\mu$) and plus or minus 2 standard deviations ($\sigma$). (i) A zoomed-in view of $V_L$ shown in (h).}
\end{figure*}

Next, we can compare the switching probability of the model. To do this, we use a sweep of $I_G$ from 0$\mu A$ to 3$\mu A$ for $I_B$ values of 14$\mu A$, 16.5, 23.5$\mu A$, 28$\mu A$, and 33$\mu A$. To evaluate the switching probability of the model, we can use the CDF evaluated at the voltage midway between the load voltage when the H-Tron is in superconducting state and when it is in resistive state. Figure \ref{fig:4} (left) shows the results of switching probability observed in the experimental data compared to the MDN model. Framing the model in this way allows us to use traditional regression performance metrics, which can be seen in Table \ref{tab:1}. On average our model is off by 0.82\% for switching probability with an $R^2$ of 0.9891. Here, the $R^2$ value represents the percentage of variance in the switching probability explained by the model.

{\renewcommand{\arraystretch}{2}
\begin{table}
\centering
\caption{\label{tab:1} Model Performance of Switching Probability }
{\setlength{\tabcolsep}{2em}
\begin{tabular}{c|c|c}
$I_B (\mu A)$ & Mean Absolute Error (MAE) & $R^2$  \\ 
\hline
14                                        & 1.149\%                            & 0.9941             \\ 
\hline
16.5                                      & 0.534\%                            & 0.9986             \\ 
\hline
23.5                                      & 1.344\%                            & 0.9572             \\ 
\hline
28                                        & 0.611\%                            & 0.9986             \\ 
\hline
33                                        & 0.662\%                            & 0.9971             \\
\hline
\end{tabular}}
\end{table}
}

Finally, we can evaluate the performance of the model in transient simulation. So that we can match the output of the model with experimental results, we mirror $I_G$ and $I_B$ from the experimental data as shown in Figure 6 (a). We use the inverse transform sampling technique as described in section II to achieve the smooth, continuous curves as seen in Figure \ref{fig:4} (right). To show the distribution of the model in the transient simulation, we report $\mu$, $\mu-2\sigma$, and $\mu+2\sigma$ conditions. We can do this by setting the quantile value ($q$) to 0.5 for the mean and 0.05 and 0.95 for the inverse transform sampling. Figure \ref{fig:4} (right) shows that our MDN model accurately captures the variance in the I-V characteristics of the devices, including the switching dynamics.

\section*{Discussion}
Our approach utilizing mixture density networks has demonstrated its remarkable ability to accurately model the variance of I-V characteristics and stochastic switching behavior of heater cryotrons. With a 0.82\% mean absolute error in the switching probability and an  $R^2$ value of 0.9891, our method has proven its efficacy in capturing the true variance of device behavior. While we've only shown the model using heater cryotrons, the generalizing power of neural networks means this approach will be easily adaptable to other devices. This achievement marks a significant stride towards more precise and reliable electronic circuit simulations, offering unprecedented opportunities for the development of cutting-edge technologies in an era where stochasticity plays an increasingly pivotal role.

\section*{Methods}

\subsection*{Heater Cryotron Fabrication}
The fabrication process began with the deposition of a 4 $nm$ layer of superconducting tungsten silicide (WSi) onto a silicon wafer through magnetron sputtering. Following this, contact pads consisting of 5 $nm$ Ti and 30 $nm$ Au were applied using optical lithography and a liftoff procedure. Subsequently, electron beam lithography was employed to define and etch the WSi layer in an RIE process with SF6 chemistry to create the device's channel. A 25 $nm$ layer of SiO$_2$ was then sputter-deposited across the entire wafer to serve as the dielectric spacer separating the heater from the channel. Another 4 $nm$ layer of WSi was subsequently sputter-deposited, patterned, and etched to form the heater input gate, with contact pads added to this upper layer through the same liftoff process as the initial layer. Finally, openings were etched through the SiO$_2$ dielectric layer using an RIE process with CHF$_3$:O$_2$ chemistry to establish electrical contact with the underlying channel layer.

\subsection*{Heater Cryotron Characterization}
The measurements in this study employed an arbitrary waveform generator (AWG) equipped with two channels and incorporated 10 $k \Omega$ resistors in series with each channel. In the data acquisition process, one channel of the AWG was configured to maintain a constant voltage, thereby delivering a fixed bias current, while the other channel was programmed to incrementally ramp up from zero current to 3 $\mu A$. Simultaneously, the voltage of the device channel was meticulously recorded on an oscilloscope. The experimental conditions covered six distinct bias current levels, spanning from 14 $\mu A$ to 33 $\mu A$. Following the ramp-up of gate current, a subsequent phase involved the concurrent reduction of gate and bias currents to get a larger breadth of data. This entire procedure was repeated 1000 times for each bias current setting, ensuring the capture of the stochastic nature of the device.

\subsection*{Model Creation}
We use Python with TensorFlow to create the mixture density network. We train the model using the data derived in the previous section. We implement a custom loss function to optimize the model for minimizing gaussian negative log-likelihood. In addition, we implement a custom activation function for exponential linear unit plus 1 plus epsilon as described in equation (3). In order to use this model in circuit simulation, we implement the neural network architecture and the inverse transform sampling methodology in Verilog-A. The trained model weights are then exported and utilized in the Verilog-A model. This process is derived from Hutchins et. al. \cite{Hutchins2022}.

\bibliography{refs}

\section*{Acknowledgements}

This research was funded by the University of Tennessee
Knoxville (https://ror.org/020f3ap87) and NIST (https://ror.org/
05xpvk416). The U.S. Government is authorized to reproduce and
distribute reprints for governmental purposes notwithstanding any copyright annotation thereon.

S. A. was supported with funds provided by the Science Alliance, a Tennessee Higher Education Commission center of excellence administered by The University of Tennessee-Oak Ridge Innovation Institute on behalf of The University of Tennessee, Knoxville.

\section*{Data Availability Statement}

The data that support the plots within this paper and other findings of this study are available from the corresponding author upon reasonable request.

\section*{Author Contributions Statement}

J.H. conceived the idea and developed the modeling framework. D.S.R and B.G.O. fabricated and characterized the heater cryotron device. A.A. supervised the project. All authors analyzed the results and contributed to writing the manuscript.  

\end{document}